%% file: IMS2014_summary_paper_template.tex
\documentclass[conference]{IEEEtran}
\IEEEoverridecommandlockouts
\usepackage{amsmath}
\usepackage{graphicx}
\usepackage{hyperref}
\usepackage{tikz}
 \usepackage{placeins} 
 \usepackage{array}
  \usepackage{caption}
  \usepackage{cite}
  \usepackage{ragged2e}
\usepackage{makecell} 
  \usepackage[caption=false,font=footnotesize]{subfig} 
\usepackage{tikz}

\usetikzlibrary{positioning}   
\usetikzlibrary{arrows.meta}   
\usetikzlibrary{fit}    

 \makeatletter
    \def\ps@IEEEtitlepagestyle{
      \def\@oddfoot{\mycopyrightnotice}
      \def\@evenfoot{}
    }
    \def\mycopyrightnotice{
      {\footnotesize 979-8-3315-7867-1/25/\$31.00~\copyright~2025 IEEE\hfill} 
     \gdef\mycopyrightnotice{}
   }

    \usepackage[mathlines,switch]{lineno}

     \makeatletter
    
    \let\old@ps@IEEEtitlepagestyle\ps@IEEEtitlepagestyle
    \def\confheader#1{%
        \def\ps@IEEEtitlepagestyle{%
            \old@ps@IEEEtitlepagestyle%
            \def\@oddhead{\strut\hfill#1\hfill\strut}%
            \def\@evenhead{\strut\hfill#1\hfill\strut}%
        }%
        \ps@headings%
    }
    \makeatother
    
    \confheader{%
            \parbox{20cm}{2025 28th International Conference on Computer and Information Technology (ICCIT)\\19-21 December, Cox’s Bazar, Bangladesh}
    }

\begin{document}

\title{\LARGE FUSE: Unifying Spectral and Semantic Cues for Robust AI-Generated Image Detection}
\twocolumn[
\begin{@twocolumnfalse}
\vfill

\fontsize{20}{24}\selectfont

\textbf{IEEE Copyright Notice}

\vspace{1em}

\fontsize{14}{17}\selectfont   

\noindent
\begin{minipage}{1.0\textwidth}
\justifying

\textcopyright\ 2025 IEEE. Personal use of this material is permitted. Permission from IEEE must be obtained for all other uses, in any current or future media, including reprinting/republishing this material for advertising or promotional purposes, creating new collective works, for resale or redistribution to servers or lists, or reuse of any copyrighted component of this work in other works. \\

\vspace{2em}

This work has been accepted for publication in \textbf{2025 28th International Conference on Computer and Information Technology (ICCIT)}. The final published version will be available via IEEE Xplore. \\

DOI: \textit{TBD}

\end{minipage}

\vfill
\end{@twocolumnfalse}
]
\author{\IEEEauthorblockN{Md. Zahid Hossain \IEEEauthorrefmark{1}, Most. Sharmin Sultana Samu \IEEEauthorrefmark{2}, Md. Kamrozzaman Bhuiyan\IEEEauthorrefmark{3}, \\Farhad Uz Zaman \IEEEauthorrefmark{4}, Md. Rakibul Islam \IEEEauthorrefmark{5}}
\IEEEauthorblockA{\IEEEauthorrefmark{1}\IEEEauthorrefmark{5}Department of CSE, Ahsanullah University of Science and Technology, Dhaka-1208, Bangladesh.\\
\IEEEauthorrefmark{2}Department of CSE, BRAC University, Dhaka-1212, Bangladesh.\\
\IEEEauthorrefmark{3}Enosis Solutions, Bangladesh\\
\IEEEauthorrefmark{4}Department of CSE, Southeast University, Dhaka-1208, Bangladesh.}
\IEEEauthorblockA{Email: \IEEEauthorrefmark{1}zahidd16@gmail.com,  \IEEEauthorrefmark{2}sharminsamu130@gmail.com,
\IEEEauthorrefmark{3}kamrozzamaan@gmail.com,\\
\IEEEauthorrefmark{4}farhad.zaman@seu.edu.bd, 
\IEEEauthorrefmark{5}rakib.aust41@gmail.com
}
}

\maketitle
\begin{abstract}
The fast evolution of generative models has heightened the demand for reliable detection of AI-generated images. To tackle this challenge, we introduce FUSE, a hybrid system that combines spectral features extracted through Fast Fourier Transform with semantic features obtained from the CLIP's Vision encoder. The features are fused into a joint representation and trained progressively in two stages. Evaluations on GenImage, WildFake, DiTFake, GPT-ImgEval and Chameleon datasets demonstrate strong generalization across multiple generators. Our FUSE (Stage 1) model demonstrates state-of-the-art results on the Chameleon benchmark. It also attains 91.36\% mean accuracy on the GenImage dataset, 88.71\% accuracy across all tested generators, and a mean Average Precision of 94.96\%. Stage 2 training further improves performance for most generators. Unlike existing methods, which often perform poorly on high-fidelity images in Chameleon, our approach maintains robustness across diverse generators. These findings highlight the benefits of integrating spectral and semantic features for generalized detection of images generated by AI.
\\

\renewcommand
\IEEEkeywordsname{Keywords}


\end{abstract}

\section{Introduction} 
The swift evolution of generative models has blurred the line between authentic and synthetic visual content. Advances in Generative Adversarial Networks (GANs) \cite{gan,biggan,gigagan,stargan,stylegan} and diffusion models \cite{diffusion,sd,vqdm,adm,glide,dalle2,openai2023dalle3,midjourney2022} have made it possible to synthesize highly realistic images that closely resemble natural ones. This development poses significant challenges in domains such as misinformation, digital forensics and creative industries, where the authenticity of visual content is critical \cite{10221755}. Consequently, robust detection of AI-generated images (AIGI) has become a pressing research problem \cite{27}. Early approaches to AIGI detection primarily employed Convolutional Neural Networks (CNNs) \cite{1, 16, 29}, which achieved strong performance by capturing pixel-level artifacts but struggled with generalization across unseen generators. More recent studies have explored Transformer-based architectures for modeling long-range dependencies and semantic cues, though these models often require substantial computational resources \cite{39,14,15,17,21,22,31}. A persistent challenge is ensuring robustness, as detectors must remain effective across diverse generative models, image resolutions and post-processing variations.

Recent advances in AIGI detection have introduced diverse strategies, including patch-based analysis \cite{13,17,22,25,31} and multimodal semantic frameworks \cite{15,19,24}. These methods achieve strong results but often rely heavily on either frequency artifacts or semantic cues. As a result, they provide incomplete representations and struggle to maintain robustness against unseen generators or post-processing. Few-shot \cite{19,20} and zero-shot \cite{21} learning have been explored to improve adaptability, yet their success remains inconsistent across different benchmarks. These limitations highlight the need for a unified framework that captures complementary information from both spectral and semantic domains.

To address this gap, we propose \textbf{FUSE}, a hybrid framework that integrates spectral and semantic features for AIGI detection. Spectral representations are obtained using Fast Fourier Transform (FFT), which captures frequency-domain regularities while reducing redundancy. Image features are obtained from the ViT encoder of a pre-trained CLIP model \cite{clip}, which captures semantic structure through joint training on large-scale vision–language data. The two feature sets are fused into a joint representation, enabling balanced and robust classification. Our framework is trained on the GenImage \cite{4genimage} and WildFake \cite{27} datasets and evaluated on both. For cross-generator testing, we further assess generalization on the DiTFake dataset (Flux, Stable Diffusion v3, PixArt-XL) \cite{31} and the GPT-ImgEval dataset (GPT-4o) \cite{34}. We also conduct experiments on the Chameleon dataset \cite{25}. Results demonstrate that the proposed approach achieves strong and consistent generalization across these challenging benchmarks. Unlike AIDE \cite{25}, which emphasizes patchwise frequency weighting through Discrete Cosine Transform, our approach integrates complementary spectral and semantic features in a unified framework to enhance robustness against diverse generators and post-processing.

The main contributions of this paper are as follows:
\begin{itemize}
\item We propose \textbf{FUSE}, a two-stage training strategy that progressively improves generalization in AIGI detection.
\item We design a spectral feature extractor and a semantic feature extractor leveraging CLIP and fuse them to form a joint representation.
\item We show that our model, trained on GenImage \cite{4genimage} and WildFake \cite{27}, achieves state-of-the-art (sota) detection performance on Chameleon dataset \cite{25} and demonstrates strong generalization to previously unseen generative models, including Diffusion Transformer-based models (Flux, Stable Diffusion v3, PixArt-XL) and GPT-4o.
\end{itemize}

The remainder of this paper is organized as follows: Section II discusses related works. Section III covers the necessary background. Section IV describes the proposed methodology. Section V reports the experimental results along with their analysis. Section VI concludes the study while highlighting potential future directions.

\section{Literature Review}
AIGI detection has progressed across multiple approaches. Early studies showed that CNN-generated images contained detectable artifacts despite high visual quality. Wang et al. \cite{30} showed ResNet-based classifiers generalized across CNN generators but struggled with diffusion models and post-processing. Later work analyzed diffusion models in the frequency domain. Corvi et al. \cite{32} found detection feasible on clean images, but performance dropped under compression and resizing, revealing robustness limits.

Large-scale benchmarks supported progress. Zhu et al. (2023) introduced GenImage, covering GAN and diffusion models for cross-generator evaluation \cite{4genimage}. Bird and Lotfi released CIFAKE, pairing 120,000 real and synthetic CIFAR-10 samples and promoting interpretability \cite{1}. Hossain et al. \cite{39} achieved 96.31\% accuracy on CIFAKE dataset with CNNs and Vision Transformers.

Wang et al. introduced DIRE, a reconstruction-based representation distinguishing diffusion models \cite{29}. Zhong et al. proposed PatchCraft, exploiting inter-pixel correlations as universal fingerprints \cite{13}.

Patch-based strategies continued. Chen et al. presented Single Simple Patch (SSP), showing small regions contained overlooked noise fingerprints \cite{17}. Epstein et al. introduced an online detection framework with release-aware sampling to adapt to evolving generators \cite{16}.

Efficiency and scalability were addressed by CLIP-based methods. Cozzolino et al. demonstrated generalization with minimal training and robustness to compression \cite{15}. Liu et al. introduced MoLE, refining CLIP-ViT with low-rank adapters while updating less than one percent of parameters \cite{24}.

Few-shot and zero-shot methods reduced training data dependence. Xu et al. proposed FAMSeC, using LoRA and contrastive learning to achieve high accuracy with thousands of samples \cite{19}. Wu et al. reframed detection as a generator-classification task in Few-Shot Detector (FSD) \cite{20}. Cozzolino et al. proposed a zero-shot entropy method modeling real image statistics \cite{21}. Karageorgiou et al. applied spectral learning using invariant frequency features for robustness across resolutions \cite{22}.

Benchmarking also advanced. Yan et al. introduced Chameleon, challenging detectors with high-resolution, realistic images \cite{25}. Hong et al. released WildFake, a 3.5-million-image dataset for systematic cross-generator evaluation \cite{27}. Li et al. proposed SAFE, an image transformation framework mitigating biases and improving local awareness \cite{31}.

These studies show rapid progress but highlight persistent challenges. Early CNN detectors lacked robustness to new models. Patch and spectral methods improved cross-model detection but required large datasets or complex adaptation. Few-shot and zero-shot methods reduced data dependence, yet performance varied. No method consistently generalizes across unseen generators.

This paper addresses these issues by combining spectral and semantic features into a joint representation and training detectors in two progressive stages. Our method reduces reliance on large datasets while maintaining adaptability to new generators.

\section{Background Study}
This section briefly reviews spectral analysis, semantic feature learning, feature fusion and evaluation metrics for AI-generated image detection.

\textbf{Spectral Analysis in Image Forensics:} Frequency-domain analysis can reveal generative artifacts using a two-dimensional Fast Fourier Transform (FFT). Patterns in magnitude and phase spectra may indicate synthesis traces. Metrics such as spectral mean and variance provide discriminative cues. Spectral features are often useful for high-fidelity generators with minimal pixel-level artifacts.

\textbf{Semantic Feature Learning:} Semantic features encode high-level visual information using the CLIP vision encoder (ViT-B/32) \cite{clip}, which is based on a Vision Transformer. The encoder divides images into patches and processes them through transformer layers with self-attention, capturing object shapes, spatial relationships and scene composition. These features can reveal content or style inconsistencies that are less detectable by low-level patterns when used with a trained classifier.

\textbf{Feature Fusion Techniques:} Spectral and semantic features are concatenated to form a joint representation. This fused representation preserves both low- and high-level representation.

\textbf{Evaluation Metrics:}
Accuracy measures the proportion of correctly classified samples. Average Precision (AP) summarizes the precision-recall trade-off for each class. AP is often more informative than accuracy for imbalanced datasets. Both metrics are used to assess generalization across multiple generators and datasets.

\section{Methodology}

\subsection{Overview}
\begin{figure}[!t]
    \centering
    \begin{tikzpicture}[node distance=2cm,>=stealth,thick]

        \node[draw, rounded corners, minimum width=2.8cm, minimum height=1cm, align=center, fill=gray!10] (input) {Input Image};

        \node[draw, rounded corners, minimum width=3.2cm, minimum height=1cm, align=center, fill=blue!10, below left=1.4cm and -0.5cm of input] (spectral) {Spectral Feature \\ Extractor (FFT)};
        
        \node[draw, rounded corners, minimum width=3.2cm, minimum height=1cm, align=center, fill=green!10, below right=1.4cm and -0.5cm of input] (semantic) {Semantic Feature \\ Extractor (CLIP)};

        \node[draw, rounded corners, minimum width=3.2cm, minimum height=1cm, align=center, fill=orange!10, below=3.2cm of input] (fusion) {Feature Fusion \\ (Concatenation)};

        \node[draw, rounded corners, minimum width=3.2cm, minimum height=1cm, align=center, fill=red!10, below=0.35cm of fusion] (classifier) {Classification Module};

        \node[draw, dashed, rounded corners, minimum width=7.5cm, minimum height=4cm, label={[align=center]above:Two-Stage Training}, fit=(spectral) (semantic) (fusion) (classifier)] (stage) {};

        \draw[->] (input.south) to[bend right=20] (spectral.north);
        \draw[->] (input.south) to[bend left=20]  (semantic.north);

        \draw[->] (spectral.south) -- ++(0,-0.3) -| (fusion.north);
        \draw[->] (semantic.south) -- ++(0,-0.3) -| (fusion.north);
        \draw[->] (fusion.south) -- (classifier.north);

    \end{tikzpicture}
    \caption{Overview of the FUSE framework.}
    \label{fig:workflow}
\end{figure}
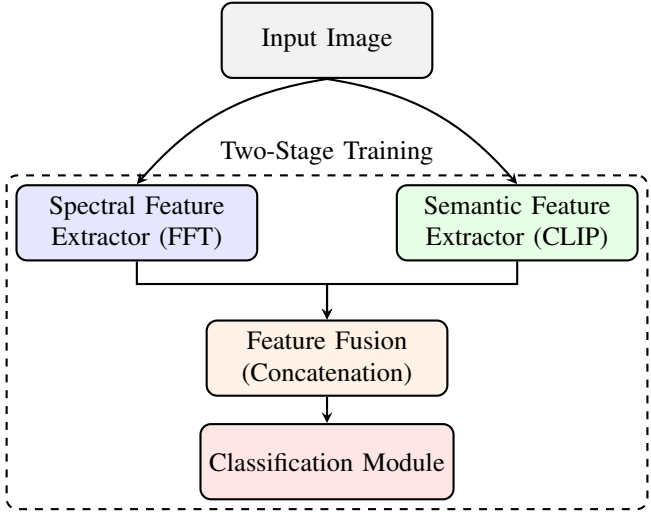

Our proposed framework leverages both spectral and semantic information to improve feature representation. The architecture consists of two main components: a Spectral Feature Extractor and a Semantic Feature Extractor, which are later combined for classification. The pipeline begins with image input, followed by spectral and semantic feature extraction. The extracted features are fused into a joint representation and passed through the classification module for prediction. The training process is conducted in two sequential stages, where the model is progressively refined to improve performance. Fig.~\ref{fig:workflow} presents the overview of the FUSE framework.

\subsection{Datasets and Training Strategy}
\begin{figure}[h]
\centerline{\includegraphics[width=0.9\columnwidth]{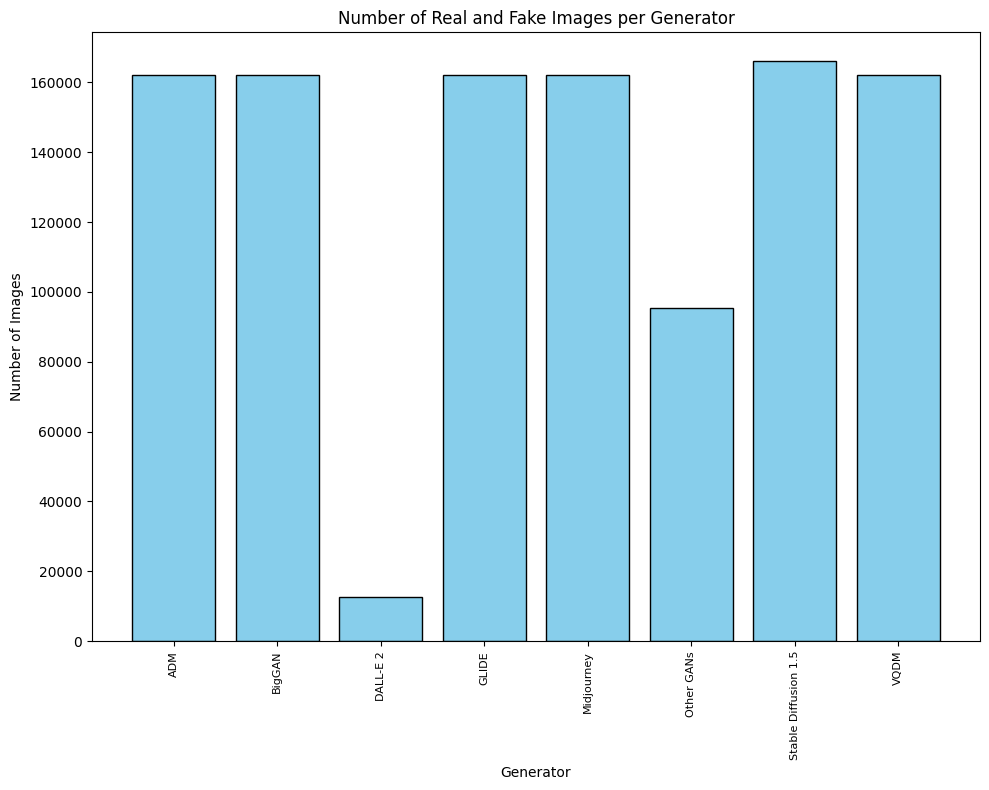}}
\caption{Number of training samples from different generators. Most samples are from GenImage \cite{4genimage}, while DALL-E 2 and other GANs (StarGAN \cite{stargan}, StyleGAN \cite{stylegan}) come from WildFake \cite{27}.}
\label{fig:train}
\end{figure}

The model is trained in two sequential stages. In the first stage, training is performed on samples from the WildFake \cite{27} and GenImage \cite{4genimage} datasets to learn initial feature representations. The number of training samples from each generator is shown in Fig.~\ref{fig:train}. Alongside the AI-generated images, 954,779 real images were used during training.  

\begin{figure}[h]
\centerline{\includegraphics[width=0.9\columnwidth]{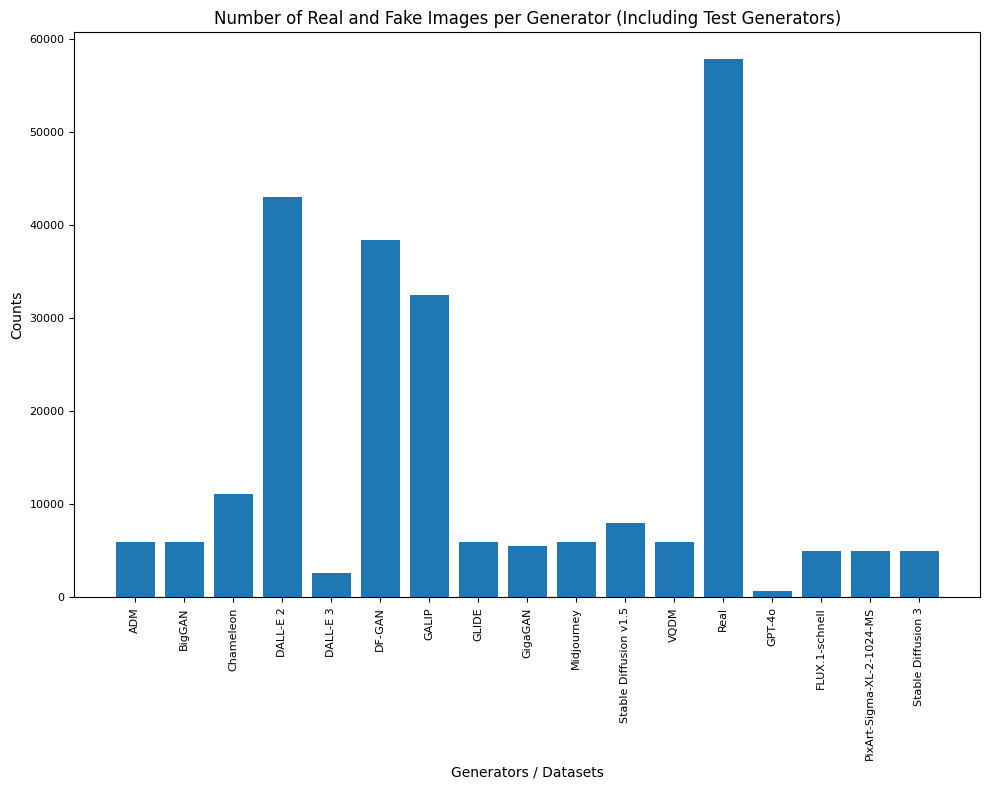}}
\caption{Number of test samples from different generators and real images. Chameleon is from Yan et al. \cite{25}, GPT-4o from GPT-ImgEval \cite{34}, and Flux, PixArt, and SD 3 from DiTFake \cite{31}. Test sets not explicitly mentioned were taken from the corresponding datasets used for first-stage training.}
\label{fig:test}
\end{figure}

In the second stage, the model is fine-tuned using 5\% of the first-stage images to mitigate catastrophic forgetting, along with additional images generated by DF-GAN, GALIP, GigaGAN, DALL-E 2 and DALL-E 3. These images were sourced from WildFake \cite{27}, with only a small subset used to adapt the learned features for improved generalization. For each generator, 5,000 real images were included as corresponding real samples. After each stage, evaluation is conducted on the test set to monitor performance and track improvements. The number of test samples for each generator is shown in Fig.~\ref{fig:test}.

\subsection{Preprocessing}
Images were standardized to a resolution of $224 \times 224$ pixels before being fed into the model. To improve robustness, an adversarial degradation module was applied during training, which randomly introduces Gaussian blur or simulated JPEG compression noise.

\subsection{Spectral Feature Extraction}
To obtain spectral features, a two-dimensional Fast Fourier Transform (FFT) was applied to the input images. The magnitude and phase spectra are computed and their mean values across spatial dimensions are extracted. This operation captures frequency-based patterns while reducing redundancy.

\subsection{Semantic Feature Extraction}
Semantic features are derived using a pre-trained CLIP model (ViT-B/32) \cite{clip}. Input images are preprocessed into CLIP’s format and encoded into high-dimensional semantic vectors. These features represent contextual and structural knowledge learned from large-scale vision-language data.

\subsection{Feature Fusion and Classification}
The outputs from both extractors are concatenated to form a joint representation. Spectral features provide complementary frequency information, while semantic features contribute contextual meaning. This fused representation is passed through the classification module for final prediction.

\subsection{Experimental Setup}
We performed our experiments on a workstation with an NVIDIA RTX 3090 GPU, an Intel 11th Gen Core i7-11700K CPU and 16 GB of RAM. The model underwent two-stage training, with 4 epochs in Stage 1 and 3 epochs in Stage 2. Optimization was performed using the Adam algorithm, with a learning rate of 0.0001 and a batch size of 64 samples.

\section{Result Analysis}

We evaluate our model after both training stages using the GenImage \cite{4genimage}, WildFake \cite{27}, DiTFake \cite{31} and GPT-ImgEval \cite{34} datasets. The model is evaluated using accuracy (Acc) and Average Precision (AP), following previous studies \cite{29,30}. For certain generators or datasets, no images belonging to the real class were available, so AP is not reported in those cases.

\begin{figure}[h]
\centerline{\includegraphics[width=1\columnwidth,height=0.3\textheight,keepaspectratio]{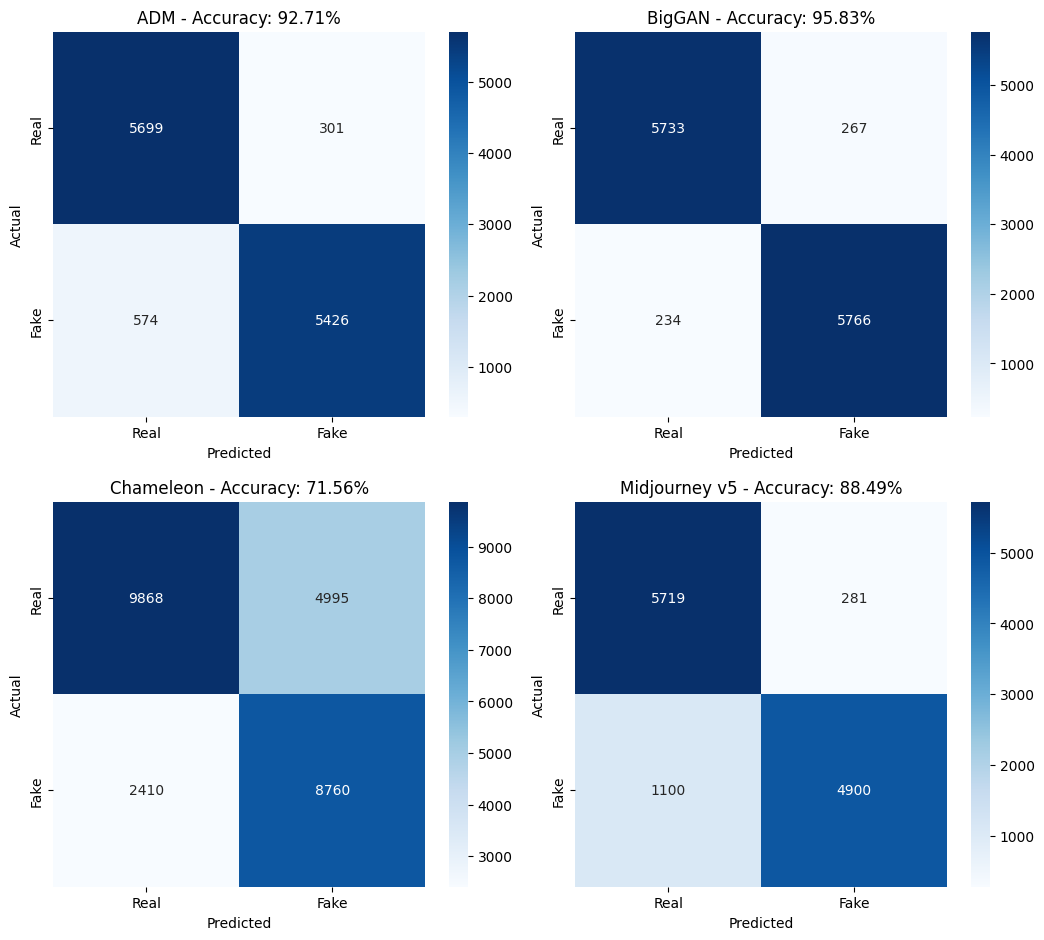}}
\caption{Confusion matrices for various generators/dataset for Stage 1.}
\label{fig7}
\end{figure}


\begin{figure}[h]
\centerline{\includegraphics[width=1\columnwidth,height=0.3\textheight,keepaspectratio]{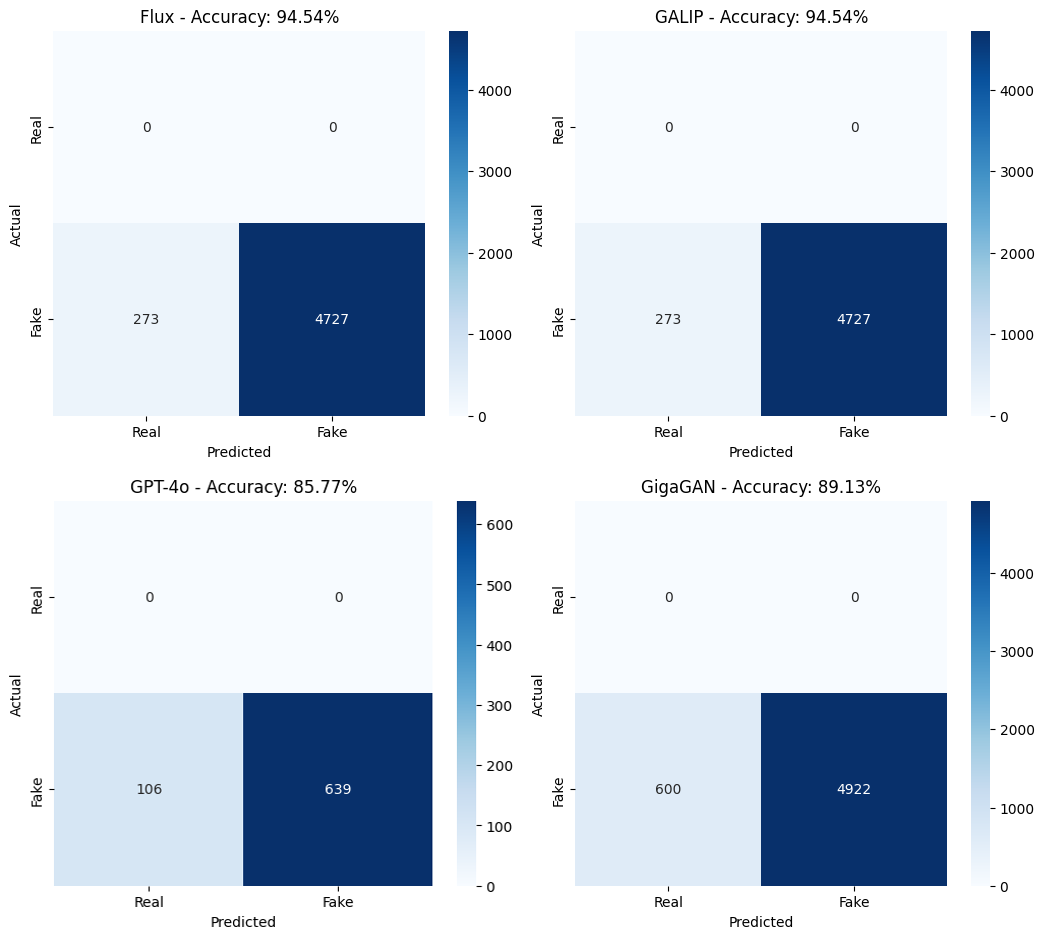}}
\caption{Confusion matrices for various generators for Stage 2.}
\label{fig5}
\end{figure}

\begin{figure}[h]
\centerline{\includegraphics[width=1\columnwidth,height=0.3\textheight,keepaspectratio]{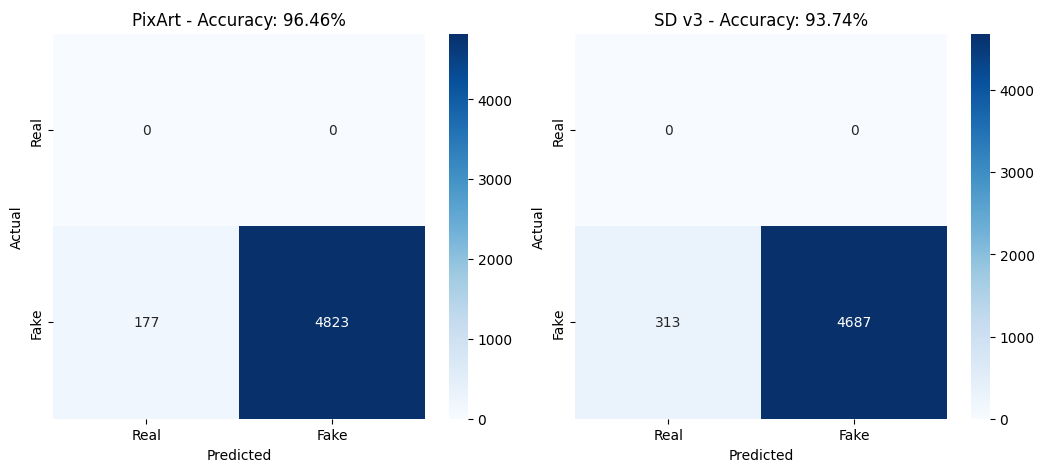}}
\caption{Confusion matrices for various generators for Stage 2.}
\label{fig6}
\end{figure}

Fig. \ref{fig7} illustrates confusion matrices of various generators and the Chameleon dataset \cite{25} for FUSE (Stage 1). The confusion matrix of Chameleon indicates that numerous real images are misclassified as fake. The model, however, performs well in identifying fake images generated by ADM, BigGAN and Midjourney.

Figures \ref{fig5} and \ref{fig6} illustrate the confusion matrices of various generators for the FUSE (Stage 2). The figures demonstrate that the model effectively detects images produced by unseen generators, including GPT-4o, SD v3, Flux and PixArt.

\begin{table*}[htbp]
\centering
\caption{Performance comparison across datasets and generators in terms of accuracy (\%). The best result for each generator is in \textbf{bold} and the second-best is \underline{underlined}. *Results were not reported in the cited paper and were obtained using the authors' models and weights. For ESSP \cite{17}, weights of the model trained with Stable Diffusion v1.4 were used. Empty cells indicate that the implementation or model weights were not publicly accessible.}
\setlength{\tabcolsep}{4pt}
\renewcommand{\arraystretch}{1.2}
\begin{tabular}{|c|c|c|c|c|c|c|c|c|c|c|c|c|c|}
\hline
Dataset $\rightarrow$
& \multicolumn{6}{c|}{GenImage} 
& \multicolumn{3}{c|}{DiTFake} 
& \multicolumn{1}{c|}{GPT4-Eval} 
& \multicolumn{1}{c|}{Chameleon} 
& \multicolumn{2}{c|}{Mean Accuracy (mAcc)} \\ 
\hline
Method $\downarrow$ &
\rotatebox{60}{ADM} &
\rotatebox{60}{BigGAN} &
\rotatebox{60}{Glide} &
\rotatebox{60}{\makecell{Midjourney \\ v5}} &
\rotatebox{60}{SD v1.5} &
\rotatebox{60}{VQDM} &
\rotatebox{60}{Flux} &
\rotatebox{60}{PixArt} &
\rotatebox{60}{SD v3} &
\rotatebox{60}{GPT-4o} &
\rotatebox{60}{\makecell{Multiple \\ Generators}} &
\rotatebox{60}{\makecell{mAcc \\(GenImage)}} &
\rotatebox{60}{mAcc (All)} \\
\hline
FUSE (Stage 1) & \textbf{92.71} & 95.83 & 92.74 & 88.49 & 92.88 & 85.53 & 90.84 & 90.46 & 91.00 & 83.76 & \textbf{71.56} & \underline{91.36} & \underline{88.71} \\
FUSE (Stage 2) & \underline{88.68} & 93.26 & 91.08 & 86.47 & 92.17 & 85.38 & \underline{94.54} & \underline{96.46} & \underline{93.74} & \underline{85.77} & 63.43 & 89.51 & 88.27 \\
AIDE \cite{25} & 78.54 & 66.89 & 91.82 & 79.38 & \textbf{99.76} & 80.26 & 85.36* & 89.89* & 81.12* & 69.70* & \underline{65.77} & 82.78 & 80.77 \\
SAFE \cite{31} & 82.10 & \underline{97.80} & \textbf{96.30} & \textbf{95.30} & \underline{99.30} & \textbf{96.30} & \textbf{99.30} & \textbf{99.60} & \textbf{99.40} & \textbf{98.92*} & 59.19* & \textbf{94.52} & \textbf{93.05} \\
PatchCraft \cite{13} & 82.17 & 95.80 & 83.79 & \underline{90.12} & 95.30 & 88.91 & -- & -- & -- & -- & -- & 89.35 & -- \\
CLIPMoLE \cite{24} & 85.70 & \textbf{98.95} & \underline{94.10} & 77.45 & 87.20 & 91.95 & -- & -- & -- & -- & -- & 89.23 & -- \\
ESSP \cite{17} & 78.90 & 73.90 & 88.90 & 82.60 & \underline{99.30} & \underline{96.00} & -- & -- & -- & -- & -- & 86.60 & -- \\
\hline
\end{tabular}
\label{tab2}
\end{table*}

\begin{table*}[htbp]
\centering
\caption{Performance comparison across datasets and generators in terms of AP (\%). *Results were not reported in the cited paper and were obtained using the authors' models and weights. Empty cells indicate that the implementation or model weights were not publicly accessible.}
\setlength{\tabcolsep}{4pt}
\renewcommand{\arraystretch}{1.2}
\begin{tabular}{|c|c|c|c|c|c|c|c|c|c|c|c|c|}
\hline
Dataset $\rightarrow$
& \multicolumn{6}{c|}{GenImage} 
& \multicolumn{3}{c|}{DiTFake} 
& \multicolumn{1}{c|}{Chameleon} 
& \multicolumn{2}{c|}{Mean Average Precision (mAP)} \\ 
\hline
Method $\downarrow$ &
\rotatebox{60}{ADM} &
\rotatebox{60}{BigGAN} &
\rotatebox{60}{Glide} &
\rotatebox{60}{\makecell{Midjourney \\ v5}} &
\rotatebox{60}{SD v1.5} &
\rotatebox{60}{VQDM} &
\rotatebox{60}{Flux} &
\rotatebox{60}{PixArt} &
\rotatebox{60}{SD v3} &
\rotatebox{60}{\makecell{Multiple\\ Generators}} &
\rotatebox{60}{\makecell{mAP\\(GenImage)}} &
\rotatebox{60}{mAP (All)} \\
\hline
FUSE (Stage 1) & \textbf{97.96} & 98.98 & \underline{97.94} & 96.31 & 97.99 & 95.56 & \underline{97.56} & \underline{97.33} & \underline{97.56} & \textbf{72.39} & \underline{97.46} & \textbf{94.96} \\
FUSE (Stage 2) & 95.47 & 98.05 & 96.65 & 94.40 & 97.25 & 94.07 & 96.53 & 96.94 & 95.80 & \underline{66.00} & 95.98 & 93.12 \\
SAFE \cite{31} & \underline{96.70} & \textbf{99.80} & \textbf{99.30} & \textbf{99.50} & \textbf{99.90} & \textbf{99.60} & \textbf{99.90} & \textbf{100.0} & \textbf{99.90} & 50.64* & \textbf{99.13} & \underline{94.52} \\
PatchCraft \cite{13} & 93.40 & \underline{99.42} & 94.04 & \underline{96.48} & \underline{99.06} & \underline{96.26} & -- & -- & -- & -- & 96.44 & -- \\
\hline
\end{tabular}
\label{tab3}
\end{table*}
Table \ref{tab2} presents a performance comparison of our methods against prior approaches in terms of accuracy (\%). Our Stage 1 model achieves sota results on ADM-generated images and the Chameleon benchmark. After Stage 2, the model shows comparable performance on ADM, FLUX, PixArt, SD v3 and GPT-4o, reflecting the benefit of progressive training. While methods such as SAFE \cite{31} outperform our models on most generators, they perform poorly on the Chameleon dataset, which contains high-fidelity images. This highlights a limitation of their approach and underscores the importance of robustness across challenging benchmarks. It should be noted that not all generators from GenImage, such as SD v1.4 and Wukong, were included in the experiments. For a fair comparison, the mean accuracy for all methods was computed only over the generators actually tested. Overall, our Stage 1 model achieves 91.36\% mean accuracy on the GenImage dataset \cite{4genimage} and 88.71\% across all tested generators, compared to 94.52\% and 93.05\% for SAFE \cite{31}, respectively. Stage 2 training improves results for several generators, including FLUX and PixArt, demonstrating that additional fine-tuning enhances cross-generator generalization.

Table \ref{tab3} presents a performance comparison of our models in terms of AP (\%) against existing approaches. FUSE (Stage 1) achieves the highest mean AP of 94.96\% across all tested generators and datasets, demonstrating balanced detection capability. For the GenImage dataset, FUSE (Stage 1) achieves 97.46\% AP, which is slightly lower than the 99.13\% achieved by SAFE \cite{31}, but still competitive across diverse generators. The AP for GPT-4o is not reported, as GPT-ImgEval does not include real images for calculation. Stage 2 training further improves performance for most generators, reflecting the benefit of additional fine-tuning. Notably, our models outperform prior methods on challenging high-fidelity images in Chameleon, highlighting robustness against complex generative outputs. These results indicate that FUSE maintains strong generalization and consistent precision across multiple datasets, generator types and training stages, while mitigating the reliance on generator-specific features.

Table \ref{tab1} reports the detection accuracy of FUSE on selected generators from the WildFake \cite{27} dataset after Stage 1 and Stage 2 training. Stage 1 results show moderate performance, with DALL-E 3 and DF-GAN achieving higher accuracy than GALIP and GigaGAN, which were completely unseen during training. DALL-E 2, partially represented in the training set, reaches 66.08\% accuracy. After Stage 2, a substantial improvement is observed across all generators, with accuracy exceeding 89\% for each, reflecting the benefits of progressive training. Overall, FUSE demonstrates strong cross-generator adaptability and robustness after two-stage training.

\begin{figure}[h]
\centerline{\includegraphics[width=1\columnwidth]{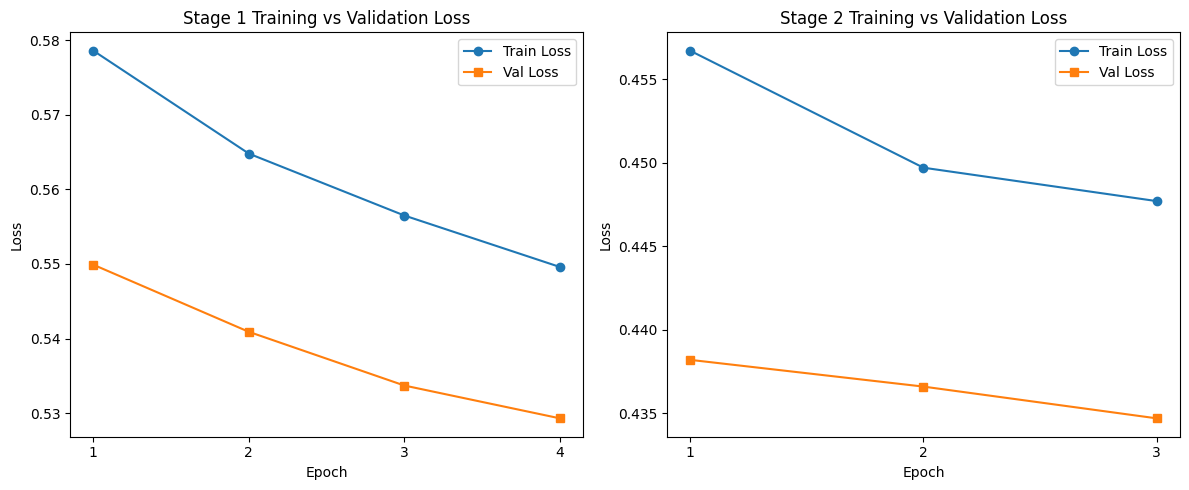}}
\caption{Training vs Validation loss in various epochs for both stages}
\label{fig8}
\end{figure}

Training and validation loss in different epochs for both stages are depicted in Fig. \ref{fig8}

\begin{table}[!htp]
\centering
\caption{Accuracy (\%) of Our Methods Across Different Generators from WildFake~\cite{27}}
\label{tab:generators}
\footnotesize
\begin{tabular}{|c|c|c|c|c|c|}
\hline
\renewcommand\cellalign{cc}
Method & 
\rotatebox{60}{DALL-E 2} & 
\rotatebox{60}{DALL-E 3} & 
\rotatebox{60}{DF-GAN} & 
\rotatebox{60}{GALIP} & 
\rotatebox{60}{GigaGAN} \\
\hline
FUSE (Stage 1) & 66.08 & 76.44 & 74.87 & 17.95 & 51.43 \\
\hline
FUSE (Stage 2) & \textbf{95.09} & \textbf{93.55} & \textbf{99.44} & \textbf{98.75} & \textbf{89.13} \\
\hline
\end{tabular}
\label{tab1}
\end{table}

\section{Conclusion and Future Work}

In this work, we proposed FUSE, a two-stage framework for detecting AI-generated images and evaluated it across multiple benchmarks, including GenImage, WildFake, DiTFake and GPT-ImgEval. Our models demonstrate strong performance in detecting both seen and unseen generators, achieving state-of-the-art accuracy and mean average precision on several datasets. Stage 2 training notably improves generalization, particularly for generators absent in the initial training stage. While our approach is effective, some high-fidelity generated images, such as those in the Chameleon dataset, remain challenging.

In future work, multimodal features and self-supervised learning strategies can be explored to further enhance robustness. Incorporating temporal cues and motion patterns could extend FUSE to AI-generated video detection. Cross-domain adaptation can be investigated to handle diverse image styles, resolutions and modalities. Exploring lightweight and efficient architectures may improve deployment on resource-constrained devices. Lastly, incorporating explainable AI methods could offer clearer understanding of the model’s predictions, enhancing trust and transparency in practical use.

\bibliographystyle{IEEEtran} 

\input{bibliography.tex}

\end{document}

%% file: bibliography.tex
\addcontentsline{toc}{chapter}{References}
\bibliography{ref}

%% file: IMS2014_summary_paper_template.bbl
\begin{thebibliography}{10}
\providecommand{\url}[1]{#1}
\csname url@samestyle\endcsname
\providecommand{\newblock}{\relax}
\providecommand{\bibinfo}[2]{#2}
\providecommand{\BIBentrySTDinterwordspacing}{\spaceskip=0pt\relax}
\providecommand{\BIBentryALTinterwordstretchfactor}{4}
\providecommand{\BIBentryALTinterwordspacing}{\spaceskip=\fontdimen2\font plus
\BIBentryALTinterwordstretchfactor\fontdimen3\font minus \fontdimen4\font\relax}
\providecommand{\BIBforeignlanguage}[2]{{%
\expandafter\ifx\csname l@#1\endcsname\relax
\typeout{** WARNING: IEEEtran.bst: No hyphenation pattern has been}%
\typeout{** loaded for the language `#1'. Using the pattern for}%
\typeout{** the default language instead.}%
\else
\language=\csname l@#1\endcsname
\fi
#2}}
\providecommand{\BIBdecl}{\relax}
\BIBdecl

\bibitem{gan}
I.~Goodfellow, J.~Pouget-Abadie, M.~Mirza, B.~Xu, D.~Warde-Farley, S.~Ozair, A.~Courville, and Y.~Bengio, ``Generative adversarial networks,'' \emph{Communications of the ACM}, vol.~63, no.~11, pp. 139--144, 2020.

\bibitem{biggan}
A.~Brock, J.~Donahue, and K.~Simonyan, ``Large scale gan training for high fidelity natural image synthesis,'' \emph{arXiv preprint arXiv:1809.11096}, 2018.

\bibitem{gigagan}
M.~Kang, J.-Y. Zhu, R.~Zhang, J.~Park, E.~Shechtman, S.~Paris, and T.~Park, ``Scaling up gans for text-to-image synthesis,'' in \emph{Proceedings of the IEEE/CVF conference on computer vision and pattern recognition}, 2023, pp. 10\,124--10\,134.

\bibitem{stargan}
Y.~Choi, M.~Choi, M.~Kim, J.-W. Ha, S.~Kim, and J.~Choo, ``Stargan: Unified generative adversarial networks for multi-domain image-to-image translation,'' in \emph{Proceedings of the IEEE conference on computer vision and pattern recognition}, 2018, pp. 8789--8797.

\bibitem{stylegan}
T.~Karras, S.~Laine, and T.~Aila, ``A style-based generator architecture for generative adversarial networks,'' in \emph{Proceedings of the IEEE/CVF conference on computer vision and pattern recognition}, 2019, pp. 4401--4410.

\bibitem{diffusion}
J.~Ho, A.~Jain, and P.~Abbeel, ``Denoising diffusion probabilistic models,'' \emph{Advances in neural information processing systems}, vol.~33, pp. 6840--6851, 2020.

\bibitem{sd}
R.~Rombach, A.~Blattmann, D.~Lorenz, P.~Esser, and B.~Ommer, ``High-resolution image synthesis with latent diffusion models,'' in \emph{Proceedings of the IEEE/CVF conference on computer vision and pattern recognition}, 2022, pp. 10\,684--10\,695.

\bibitem{vqdm}
S.~Gu, D.~Chen, J.~Bao, F.~Wen, B.~Zhang, D.~Chen, L.~Yuan, and B.~Guo, ``Vector quantized diffusion model for text-to-image synthesis,'' in \emph{Proceedings of the IEEE/CVF conference on computer vision and pattern recognition}, 2022, pp. 10\,696--10\,706.

\bibitem{adm}
P.~Dhariwal and A.~Nichol, ``Diffusion models beat gans on image synthesis,'' \emph{Advances in neural information processing systems}, vol.~34, pp. 8780--8794, 2021.

\bibitem{glide}
A.~Nichol, P.~Dhariwal, A.~Ramesh, P.~Shyam, P.~Mishkin, B.~McGrew, I.~Sutskever, and M.~Chen, ``Glide: Towards photorealistic image generation and editing with text-guided diffusion models,'' \emph{arXiv preprint arXiv:2112.10741}, 2021.

\bibitem{dalle2}
A.~Ramesh, P.~Dhariwal, A.~Nichol, C.~Chu, and M.~Chen, ``Hierarchical text-conditional image generation with clip latents,'' \emph{arXiv preprint arXiv:2204.06125}, 2022.

\bibitem{openai2023dalle3}
OpenAI, ``Dall·e 3,'' \url{https://openai.com/dall-e-3}, 2023, accessed on May 14, 2025.

\bibitem{midjourney2022}
``Midjourney,'' \url{https://www.midjourney.com/}, 2022, accessed on May 14, 2025.

\bibitem{10221755}
Y.~Wang, Y.~Pan, M.~Yan, Z.~Su, and T.~H. Luan, ``A survey on chatgpt: Ai--generated contents, challenges, and solutions,'' \emph{IEEE Open Journal of the Computer Society}, vol.~4, pp. 280--302, 2023.

\bibitem{27}
Y.~Hong, J.~Feng, H.~Chen, J.~Lan, H.~Zhu, W.~Wang, and J.~Zhang, ``Wildfake: A large-scale and hierarchical dataset for ai-generated images detection,'' in \emph{Proceedings of the AAAI Conference on Artificial Intelligence}, vol.~39, no.~4, 2025, pp. 3500--3508.

\bibitem{1}
J.~J. Bird and A.~Lotfi, ``Cifake: Image classification and explainable identification of ai-generated synthetic images,'' \emph{IEEE Access}, vol.~12, pp. 15\,642--15\,650, 2024.

\bibitem{16}
D.~C. Epstein, I.~Jain, O.~Wang, and R.~Zhang, ``Online detection of ai-generated images,'' in \emph{2023 IEEE/CVF International Conference on Computer Vision Workshops (ICCVW)}, 2023, pp. 382--392.

\bibitem{29}
Z.~Wang, J.~Bao, W.~Zhou, W.~Wang, H.~Hu, H.~Chen, and H.~Li, ``Dire for diffusion-generated image detection,'' in \emph{Proceedings of the IEEE/CVF International Conference on Computer Vision (ICCV)}, October 2023, pp. 22\,445--22\,455.

\bibitem{39}
M.~Z. Hossain, F.~U. Zaman, and M.~R. Islam, ``Advancing ai-generated image detection: Enhanced accuracy through cnn and vision transformer models with explainable ai insights,'' in \emph{2023 26th International Conference on Computer and Information Technology (ICCIT)}.\hskip 1em plus 0.5em minus 0.4em\relax IEEE, 2023, pp. 1--6.

\bibitem{14}
Z.~Xi, W.~Huang, K.~Wei, W.~Luo, and P.~Zheng, ``Ai-generated image detection using a cross-attention enhanced dual-stream network,'' in \emph{2023 Asia Pacific Signal and Information Processing Association Annual Summit and Conference (APSIPA ASC)}.\hskip 1em plus 0.5em minus 0.4em\relax IEEE, 2023, pp. 1463--1470.

\bibitem{15}
D.~Cozzolino, G.~Poggi, R.~Corvi, M.~Nie{\ss}ner, and L.~Verdoliva, ``Raising the bar of ai-generated image detection with clip,'' in \emph{Proceedings of the IEEE/CVF Conference on Computer Vision and Pattern Recognition}, 2024, pp. 4356--4366.

\bibitem{17}
J.~Chen, J.~Yao, and L.~Niu, ``A single simple patch is all you need for ai-generated image detection,'' \emph{arXiv preprint arXiv:2402.01123}, 2024.

\bibitem{21}
D.~Cozzolino, G.~Poggi, M.~Nie{\ss}ner, and L.~Verdoliva, ``Zero-shot detection of ai-generated images,'' in \emph{European Conference on Computer Vision}.\hskip 1em plus 0.5em minus 0.4em\relax Springer, 2024, pp. 54--72.

\bibitem{22}
D.~Karageorgiou, S.~Papadopoulos, I.~Kompatsiaris, and E.~Gavves, ``Any-resolution ai-generated image detection by spectral learning,'' in \emph{Proceedings of the Computer Vision and Pattern Recognition Conference}, 2025, pp. 18\,706--18\,717.

\bibitem{31}
O.~Li, J.~Cai, Y.~Hao, X.~Jiang, Y.~Hu, and F.~Feng, ``Improving synthetic image detection towards generalization: An image transformation perspective,'' in \emph{Proceedings of the 31st ACM SIGKDD Conference on Knowledge Discovery and Data Mining V. 1}, 2025, pp. 2405--2414.

\bibitem{13}
N.~Zhong, Y.~Xu, S.~Li, Z.~Qian, and X.~Zhang, ``Patchcraft: Exploring texture patch for efficient ai-generated image detection,'' \emph{arXiv preprint arXiv:2311.12397}, 2023.

\bibitem{25}
S.~Yan, O.~Li, J.~Cai, Y.~Hao, X.~Jiang, Y.~Hu, and W.~Xie, ``A sanity check for ai-generated image detection,'' \emph{arXiv preprint arXiv:2406.19435}, 2024.

\bibitem{19}
J.~Xu, Y.~Yang, H.~Fang, H.~Liu, and W.~Zhang, ``Famsec: A few-shot-sample-based general ai-generated image detection method,'' \emph{IEEE Signal Processing Letters}, 2024.

\bibitem{24}
Z.~Liu, H.~Wang, Y.~Kang, and S.~Wang, ``Mixture of low-rank experts for transferable ai-generated image detection,'' \emph{arXiv preprint arXiv:2404.04883}, 2024.

\bibitem{20}
S.~Wu, J.~Liu, J.~Li, and Y.~Wang, ``Few-shot learner generalizes across ai-generated image detection,'' \emph{arXiv preprint arXiv:2501.08763}, 2025.

\bibitem{clip}
A.~Radford, J.~W. Kim, C.~Hallacy, A.~Ramesh, G.~Goh, S.~Agarwal, G.~Sastry, A.~Askell, P.~Mishkin, J.~Clark \emph{et~al.}, ``Learning transferable visual models from natural language supervision,'' in \emph{International conference on machine learning}.\hskip 1em plus 0.5em minus 0.4em\relax PmLR, 2021, pp. 8748--8763.

\bibitem{4genimage}
M.~Zhu, H.~Chen, Q.~Yan, X.~Huang, G.~Lin, W.~Li, Z.~Tu, H.~Hu, J.~Hu, and Y.~Wang, ``Genimage: A million-scale benchmark for detecting ai-generated image,'' \emph{Advances in Neural Information Processing Systems}, vol.~36, pp. 77\,771--77\,782, 2023.

\bibitem{34}
Z.~Yan, J.~Ye, W.~Li, Z.~Huang, S.~Yuan, X.~He, K.~Lin, J.~He, C.~He, and L.~Yuan, ``Gpt-imgeval: A comprehensive benchmark for diagnosing gpt4o in image generation,'' \emph{arXiv preprint arXiv:2504.02782}, 2025.

\bibitem{30}
S.-Y. Wang, O.~Wang, R.~Zhang, A.~Owens, and A.~A. Efros, ``Cnn-generated images are surprisingly easy to spot... for now,'' in \emph{Proceedings of the IEEE/CVF conference on computer vision and pattern recognition}, 2020, pp. 8695--8704.

\bibitem{32}
R.~Corvi, D.~Cozzolino, G.~Zingarini, G.~Poggi, K.~Nagano, and L.~Verdoliva, ``On the detection of synthetic images generated by diffusion models,'' in \emph{ICASSP 2023-2023 IEEE International Conference on Acoustics, Speech and Signal Processing (ICASSP)}.\hskip 1em plus 0.5em minus 0.4em\relax IEEE, 2023, pp. 1--5.

\end{thebibliography}
